# A Machine Learning Approach for Predicting Deterioration in Alzheimer's Disease

Henry Musto, Daniel Stamate, Ida Pu
Data Science and Soft Computing Lab, and
Department of Computing
*Goldsmith, University of London*
London, United Kingdom
Email: hthom018@gold.ac.uk

Daniel Stahl
Department of Biostatistics and Health Informatics
Institute of Psychiatry, Psychology, and Neuroscience
*Kings College London*
London, United Kingdom

*Abstract* - **This paper explores deterioration in Alzheimer's Disease using Machine Learning. Subjects were split into two datasets based on baseline diagnosis (Cognitively Normal, Mild Cognitive Impairment), with outcome of deterioration at final visit (a binomial essentially yes/no categorisation) using data from the Alzheimer's Disease Neuroimaging Initiative (demographics, genetics, CSF, imaging, and neuropsychological testing etc). Six machine learning models, including gradient boosting, were built, and evaluated on these datasets using a nested cross-validation procedure, with the best performing models being put through repeated nested cross-validation at 100 iterations. We were able to demonstrate good predictive ability using CART predicting which of those in the cognitively normal group deteriorated and received a worse diagnosis (AUC = 0.88). For the mild cognitive impairment group, we were able to achieve good predictive ability for deterioration with Elastic Net (AUC = 0.76).**

## I. INTRODUCTION

Alzheimer's Disease which accounts for 60-80% of all dementias [1], remains a pernicious threat to older people. As of 2017, it is estimated that over 50 million people globally are living with dementia [2]. Unlike other chronic diseases such as cancer and heart disease, dementia can require intensive care for many years after diagnosis. As a result, the cost of care is significantly higher as sufferers increasingly require round-the-clock care. Furthermore, the burden of caring for someone with dementia can negatively impact the wider families wellbeing, with associated detrimental effects on mental and physical health. The cost of long-term care also puts a disproportionate financial burden onto families and the wider community of those with a diagnosis. It is estimated that the cost of caring for those with dementia currently stands at 1% of global GDP [3]. Furthermore, 60% of those with dementia live in low- and middle-income countries [4] and, as these countries transition into developed economies, with the associated lengthening life expectancy and falling birth rate, the strain on these nations to care for dementia sufferers will become financially untenable.

In lieu of effective treatment for Alzheimer's disease [5], research has turned towards the possibility of early detection of Alzheimer's biomarkers before the onset of symptoms [6]. Work in this area has noted that accurate prediction of the risk of developing Alzheimer's disease and subsequent early interventions aimed at delaying the onset of symptoms would ease the burden of suffering as well as financial costs of care for the patient and their family. Even if no intervention were available, a pre-warning of the risk of dementia would allow patients and carers to prepare for the possibility of cognitive decline, and thus may help reduce the psychological and practical effects of diagnosis. Indeed, research has suggested that the structural brain changes that precipitate Alzheimer's symptoms may begin several years before the onset of notable symptoms [6], and this may provide an opportunity to develop techniques for assessing dementia risk.

There have been several attempts to create predictive models for Alzheimer's Disease. Mathotaarachchiet et al. used the same repository (ADNI) as the present paper to build Support Vector Machine (SVM) and Logistic Regression models with PET neuroimages for SVM and regularized logistic regression, achieving an AUC of 0.91 [7]. Casanova et al. investigated a range of models as they explored genetic and non-genetics data as predictors of cognitive decline. The best model found in that study was Random Forest with an accuracy of 78% [8]. The work of Stamate et al used several different machine learning techniques in the exploration of the ADNI dataset, with a view to assessing the predictive power of variables found in the dataset. This work utilised, among others, the Gaussian Process technique which has, hitherto, seldom been explored in this context [9]. The present paper also explores several techniques including Gaussian Process. Gill et al. collapsed the final visit diagnosis of baseline cognitively normal subjects, into a binary classification indicating whether the subject had deteriorated and received a diagnosis of *either* Mild Cognitive Impairment (MCI) or Alzheimer's Dementia (AD). The model from Gill et al. achieved a good predictive power within this paradigm (AUC 84.4%) [10].

This paper emulates this binary classification for cognitively normal subjects, but also extends it for those who were diagnosed with mild cognitive impairment at baseline. We seek to utilise statistical learning as a mechanism to predict those who would suffer cognitive deterioration resulting in receiving a diagnosis of Mild Cognitive Impairment *or* Alzheimer's Disease. To this end, we will split the well-known dataset from the Alzheimer's Disease Neuroimaging Initiative (ADNI) into two separate groups at baseline: those who were Cognitively Normal (CN), and those who received a diagnosis of Mild Cognitive Impairment (MCI). We then predict what diagnosis each subject received upon their final visit to the ADNI test sites. We tune and test six different models on these datasets separately. As discussed, this study collapses the multinomial diagnosis received at final visit into a binomial progression outcome of whether there is a deterioration, i.e., received either "no deterioration": the same diagnosis (or a more favourable one) or deteriorated and consequently received a worse diagnosis at final visit. Such a separation would have potential clinical benefits, as it would allow a greater understanding of the mechanisms that underpin deterioration for the two groups. An accurate prediction would serve as both a predictive and inferential tool as we may be afforded greater understanding of potentially modifiable risk factors that, if interventions were implemented, would allow delay or prevention of the onset of cognitive decline. Thus, this paper aims to join an existing body of work which specifically looks at dementia prediction through the lens of deterioration across a longitudinal data collection period.

The goal of this study is therefore to predict, using predictors derived at baseline, those who would go on to receive a *worse* diagnosis upon their last visit to a testing site.

## II. METHODS

### A. *Alzheimer's Disease Neuroimaging Initiative.*

The Data used in this paper was derived from the Alzheimer's Disease Neuroimaging Initiative (ADNI) database. This multicentre longitudinal study was initiated in 2004 by the National Institute of Aging (NIA), The National Institute of Biomedical Imaging and Bioengineering (NIBIB), The Food and Drug Administration (FDA), private pharmaceutical companies and non-profit organizations in the US. It was conducted over six years, with 400 subjects diagnosed with Mild Cognitive Impairment, 200 subjects with Alzheimer's Disease and 200 healthy controls. The initial goal of the ADNI study was to test whether repeated collections of neuroimaging, biomarker, genetic, and clinical and neuropsychological data could be combined to contribute in an impactful way to research into dementia [11].

Data for the present paper was downloaded on the 10th of June 2021 through the ADNIMERGE package in R. This package combines several variables from the different ADNI datasets and studies (ADNI1, ADNIGO, ADNI2, and ADNI3). The final combined dataset contains 115 variables and 15,157 observations, which included multiple observations per participant. These observations represent data collection events, where participants made multiple (up to 23) visits to study sites.

The data used for this paper is a subset of the full dataset, containing only information from the original ADNI1 study. The different ADNI protocols contain diagnostic classifications that have been reached using procedures that differ across the studies. Thus, an attempt to classify based on diagnostic labels in ADNI1 will necessitate different predictors than in ADNI2, for example. Therefore, the decision was taken to use only data from ADNI1. After some initial cleaning, the resulting data for this paper contains 35 variables with 5,013 observations, with 34 input attributes, and 1 outcome attribute. The outcome attribute consisted of three diagnostic classes: those who received a diagnosis of Cognitively Normal (**CN**), those who received a diagnosis of Mild Cognitive Impairment (**MCI**), and those who received a diagnosis of Alzheimer's Disease (**AD**).

### B. *Description of Variables*

- **Baselines Demographics:** age, gender, ethnicity, race, marital status, and education level were included in the original dataset. All were preserved after pre-processing, but nominal values were dummified to numeric format.

- **Functional Activities Questionnaire (FAQ)** is a test that can be used to assess the dependency on another person that a participant requires to carry out normal daily tasks. FAQ consists of questionnaires and multiple-choice questions, which are given to yield an aggregate score from 0 to 30.

- **Mini-Mental State Exam (MMSE)** is used to estimate the severity and progression of cognitive impairment and to follow the course of cognitive changes in an individual over time.

- **PET measurements (FDG, PIB, AV45)** are indirect measures of brain function.

- **MRI measurements (Hippocampus, intracranial volume (ICV), MidTemp, Fusiform, Ventricles, Entorhinal and WholeBrain)** are structural measurements of a participant's brain.

- **APOE4** is an integer measurement representing the appearance of epsilon 4 allele of the APOE gene.

- **Variables 'ABETA', 'TAU', 'PTAU'** are cerebrospinal fluid (CSF) biomarker measurements.

- **Rey's Auditory Verbal Learning Test (RAVLT)** are neurophysiological test evaluating an individual's episodic memory.

- **Everyday cognitive evaluations (Ecog)** are questionnaires that illustrates a participant's ability to carry out everyday tasks.

- **Logical Memory – Delayed Recall Total Number of Story Units Recalled (LDELTOTAL)** is a

neuropsychological test that evaluates a person's ability to recall information after a prescribed amount of time.

- **Modified Preclinical Alzheimer Cognitive Composite (mPACC)** are tests that evaluate a person's cognition, episodic memory and timed executive function.

- **ADAS** and **MOCA** are generalized neuropsychological tests that evaluate a person's cognitive ability (e.g., memory, visuospatial, etc.).

- **Last Visit** is a variable, defined for this paper as the number of months from baseline data collection to the subject's last visit at a test centre. This variable was added to control for the differing time periods between subjects first and last visits.

### C. Data Cleaning and Missing Value Imputation

The data was cleaned using the caret package in R. The *preProcess* function was used to implement a centring and scaling of the data. KNN was used with K = 5 to estimate and replace missing data. However, predictors with more than 90% missing data were excluded from the final dataset. Thus, the variable PIB was excluded. All nominal predictors were dummy coded. The final dataset, therefore, contained 42 predictor variables and 1 outcome variable. The pre-processing was conducted within the nested cross-validation procedure described below, with the training and test data being pre-processed separately to avoid any 'information bleed' that could bias the model's performance measures, with the pre-processing model trained on the training set used to pre-process the test dataset.

### D. Diagnostic Classes

The goal of this study was to predict, using predictors derived at baseline, those who would go on to receive a *worse* diagnosis upon their last visit to a testing site. Therefore, the data was split into two groups, denoting the different diagnoses received at baseline.

Those who received a diagnosis of **CN** at baseline were in group one (CN_b group), with the goal of predicting whether they received the *same diagnosis at their last visit or received a worse diagnosis*. For the sake of ease of inference, the two worse diagnoses (MCI and AD) were combined into a single classification, **'MCI/AD'**.

Those who received a diagnosis of **MCI** at baseline were in the second group (MCI_b group), with the goal of predicting whether they received the *same diagnosis/received a more favourable diagnosis at their last visit or received a worse diagnosis*. For the sake of ease of inference, subjects who received a diagnosis of **MCI** or **CN** were combined into a single classification (**'CN/MCI'**). The worse diagnosis was denoted by the classification labels of **AD.**

TABLE I. THOSE WHO RECEIVED A DIAGNOSIS OF COGNITIVELY NORMAL (CN) AT BASELINE WERE THE ONLY GROUP INCLUDED. THE MODELS PREDICTED WHAT DIAGNOSIS WERE RECEIVED FOR THESE SUBJECTS AT THE FINAL VISIT, WITH THE BINOMIAL OUTCOMES DETAILED HERE.

| Outcome | Definition |
|---|---|
| **Cognitively Normal (CN)** | Those, having received a diagnosis of CN at baseline, received the same diagnosis at their last visit. |
| **Mild Cognitive Impairment (MCI/AD)** | Those, having received a diagnosis of CN at baseline, *either* received a diagnosis of AD or MCI at their last visit. |

TABLE II. THOSE WHO RECEIVED A DIAGNOSIS OF MILD COGNITIVE IMPAIREMENT (MCI) AT BASELINE WERE THE ONLY GROUP INCLUDED. THE MODELS PREDICTED WHAT DIAGNOSIS WERE RECEIVED FOR THESE SUBJECTS AT THE FINAL VISIT, WITH THE BINOMIAL OUTCOMES DETAILED HERE.

| Classification | Definition |
|---|---|
| **Cognitively Normal/Mild Cognitive Impairment (CN/MCI)** | Those, having received a diagnosis of MCI at baseline, *either* received the same diagnosis at their last visit or received a more favourable diagnosis of CN. |
| **Alzheimer's Disease (AD)** | Those, having received a diagnosis of MCI at baseline, received a diagnosis of AD at their last visit. |

The resulting datasets had the following dimensions:

TABLE III. THE FINAL DIMENSIONS OF THE TWO DATASETS AFTER PREPROCESSING.

| Dataset | Variables | Observations/Subjects |
|---|---|---|
| **CN at baseline** | 43 | 285 |
| **MCI at baseline** | 43 | 392 |

### E. Model Tuning

To produce optimised predictive models, we controlled the parameter values for each algorithm using grid searches. Models were fitted with a 5-fold cross-validation, after being pre-processed, using the procedures described above. Models were then evaluated on the test sets. We chose a range of models to train using the ADNI data. The models trained used the following algorithms:

1. Random Forest (RF)
2. Support Vector Machines with a radial kernel (SVM)
3. Gradient Boosting Machine (GBM)
4. Elastic Net (EN)
5. Gaussian Processes with a radial kernel (GP)
6. Classification and Regression Tree (CART)

The models were tuned with the following range of values for each dataset.

- CN_b Group

Random Forest models were tuned over mtry between 1 and the number of columns for the training dataset, max depth between 1-10, minimum split improvement between 0.01 and 0.2 by intervals of 0.01, minimum rows between 1-7. The optimal values were found at mtry = 6, max depth = 9, minimum rows = 1, and minimum split improvement = 0.06.

Support Vector Machines with radial kernels were tuned with C from 0 to 5 by intervals of 0.1, and sigma from 0 to 5, by intervals of 0.1. The optimal values were found at sigma = 0.8, and C = 0.2.

Gradient Boosting Machine models were tuned with trees from 1 to 400, max depth values of 25-100 by intervals of 25, learn rate from 0.01 to 0.2 by 0.1. and minimum rows between 1-50. The optimal values were found at tree = 10, max depth = 50, min rows = 40 and minimum split improvement = 0.00001.

Elastic Net was tuned over values of lambda from 0 to 10 by 0.1 and alpha from 0 to 1 by 0.01. the optimal values were found at lambda =10, alpha = 0.

Gaussian Process with a radial kernel was tuned with sigma from 0.001- 2 by intervals of 0.001. The optimal value for sigma was found at 1.288.

CART was tuned over complexity between 1-250 with whole number values. The optimal value was cp = 200.

- MCI_b Group

Random Forest models were tuned over mtry between 1 and the number of columns for the training dataset, max depth between 1-10, minimum split improvement between 0.01 and 0.2 by intervals of 0.01, minimum rows between 1-7. The optimal values were found at mtry = 6, max depth = 10, minimum rows = 4, and minimum split improvement = 0.06.

Support vector machines with radial kernels were tuned with C from 0 to 5 by intervals of 0.1, and sigma from 0, 5, by intervals of 0.1. The optimal values were found at C = 0.1, Sigma = 0.1.

Gradient Boosting Machine models were tuned with trees from 1 to 400, max depth values of 25-100 by intervals of 25, learn rate from 0.01 to 0.2 by 0.1. and minimum rows between 1-50. The optimal values were found at tree = 100, max depth = 50, min rows = 40 and minimum split improvement = 0.00001.

Elastic Net was tuned over values of lambda from 0 to 10 by 0.1 and alpha from 0 to 1 by 0.01. the optimal values were found at lambda =3, alpha = 0.01.

Gaussian Process with a radial kernel was tuned with sigma from 0.001- 2 by intervals of 0.001. The optimal value for sigma was found at 1.191.

CART was tuned over complexity between 1-250 with whole number values. The optimal value was cp = 200.

### F. Nested cross-validation

The CN_b group suffered from a lack of data points. Once train/test splitting was performed, the number of rows in the test data was less than 100. This led to very high variances within the models. Thus, it was decided to implement a form of nested cross-validation to estimate the performance of a large test set. This procedure involves randomly creating N-folds, where N = number of rows/3. For each fold, three randomly selected rows were taken for the test set, with the remainder as the training dataset. On the training dataset, a 5-fold cross-validation is used, where we tune different hyperparameters for the model. Once the optimal hyperparameters are found, and the model built, we predict on the test set, and record the probabilities that are returned. We then proceed onto the next fold where the process is repeated. This continues until all folds have been used, resulting in all data points being predicted as part of the test set. This results in a vector of probabilities, one for each row of the complete dataset. From there we can then compute the performance statistics. We can then proceed to test these statistics' stability using the repeated nested cross-validation procedure described below. For the sake of consistency, we ran this procedure on both the CN and MCI datasets. Further post-processing was performed on summary statistics, taking the sensitivity, accuracy, specificity, and kappa statistics at the Youden point on the ROC curve.

### G. Repeated nested cross-validation procedure.

The stability of a best performing subset of the models was explored using a repeated nested cross-validation procedure. This procedure involves running the nested cross-validation procedure 100 times, such that we can demonstrate the variance in model performance. For this simulation we repeated the process 100 times, and the mean and standard deviations of the AUC was taken. Only one model from the nested CV procedure from each dataset was chosen for this extension.

### III. RESULTS

The top performing model for the CN data was based on the CART with an AUC of 0.88. This indicates a significantly strong pattern and decision boundary in the data. The CART nested cross-validation was then run 100 times to test stability, with a resulting average AUC of 0.89 (SD = 0.006). When considering the variable importance chart for this model we can see neuropsychological testing being the most important predictors.

The top performing model for the MCI data was based on the Elastic Net with an AUC of 0.75. This indicates a significantly strong pattern and decision boundary in the data. The Elastic Net, after going through the repeated nested CV at 100 iterations achieved an AUC of 0.76 (SD = 0.01). When considering variable importance, the 'last_visit' predictor proved to be the most important when building the best performing model. However, unlike the CN dataset, some MRI imaging measurements; the midtempal and hippocampal volume, proved to be important, with less neuropsychological tests appearing in the top important predictors in this model.

TABLE IV. STATISTICS FOR MODELS APPLIED TO THE CN_b GROUP.

| Model | AUC | Sensitivity | Specificity | Accuracy | Kappa |
|---|---|---|---|---|---|
| Random Forest | 0.83 | 0.72 | 0.89 | 0.73 | 0.24 |
| SVM with Radial Kernel | 0.85 | 0.77 | 0.94 | 0.79 | 0.33 |
| GBM | 0.87 | 0.80 | 0.94 | 0.82 | 0.37 |
| Elastic Net | 0.60 | 0.65 | 0.11 | 0.92 | 0.15 |
| Gaussian Process | 0.59 | 0.33 | 0.86 | 0.82 | 0.14 |
| CART | 0.88 | 0.76 | 1 | 0.78 | 0.33 |

TABLE V. STATISTICS FOR MODELS APPLIED TO THE MCI_b DATASET.

| Model | AUC | Sensitivity | Specificity | Accuracy | Kappa |
|---|---|---|---|---|---|
| Random Forest | 0.67 | 0.91 | 0.24 | 0.53 | 0.13 |
| SVM with radial kernel | 0.69 | 0.35 | 0.82 | 0.62 | 0.18 |
| GBM | 0.62 | 0.82 | 0.37 | 0.56 | 0.17 |
| Elastic Net | 0.75 | 0.80 | 0.60 | 0.68 | 0.38 |
| Gaussian Process | 0.66 | 0.88 | 0.47 | 0.65 | 0.33 |
| CART | 0.62 | 0.60 | 0.60 | 0.60 | 0.20 |

TABLE VI. MEAN(SD) VALUES OF PERFORMANCE STATISTICS FOR REPEATED NESTED CROSSVALIDATION OF THE CART MODEL FOR THE CN GROUP.

| AUC | 0.89(0.006) |
|---|---|
| Sensitivity | 0.76 (0.008) |
| Specificity | 1(0) |
| Accuracy | 0.78(0.007) |
| Kappa | 0.34(0.01) |

TABLE VII. MEAN(SD) VALUES OF PERFORMANCE STATISTICS FOR REPEATED NESTED CROSSVALIDATION OF THE ELASTIC NET MODEL FOR THE MCI GROUP.

| AUC | 0.76(0.01) |
|---|---|
| Sensitivity | 0.68 (0.06) |
| Specificity | 0.76(0.06) |
| Accuracy | 0.71(0.01) |
| Kappa | 0.43(0.02) |

## IV. Discussion and Conclusions

This paper represents a broad attempt to classify on an existing dataset, using a range of techniques. We were able to demonstrate that, by collapsing final diagnoses into a binary classification problem, one can achieve good results using several recognised algorithms. We further demonstrated marked differences in the apparent decision boundaries, when predicting for those with a diagnosis of Cognitively Normal, vs Mild Cognitive Impairment, at baseline. The separation of the two groups in this study was required, as the outcome of the CN_b group shared the same diagnoses as the baseline of the second group, however this did afford us the opportunity to study these groups as distinct datasets. Such findings would indicate that there is value in treating these two groups separately. To be more explicit, the task of predicting which cognitively normal elderly people will deteriorate, is apparently different from predicting which of those with subjective memory complaints will get worse, with differing predictors of import. Such a finding should inform future studies, as we move away from 'one size fits all' approach to ML as a predictive tool within dementia research.

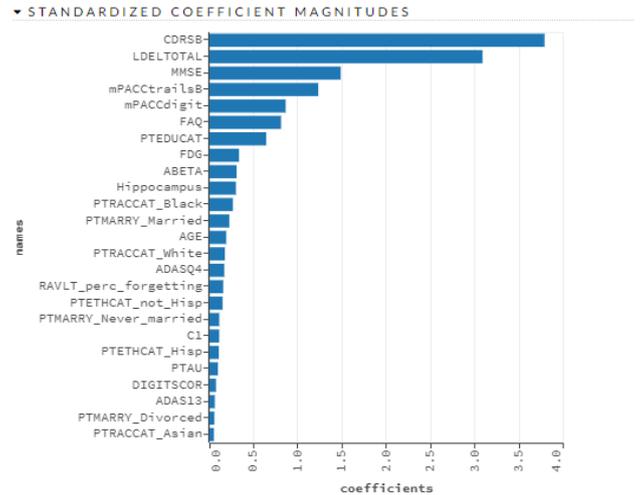

Fig. 1. Variable Importance for the CN_b group from a Random Forest model.

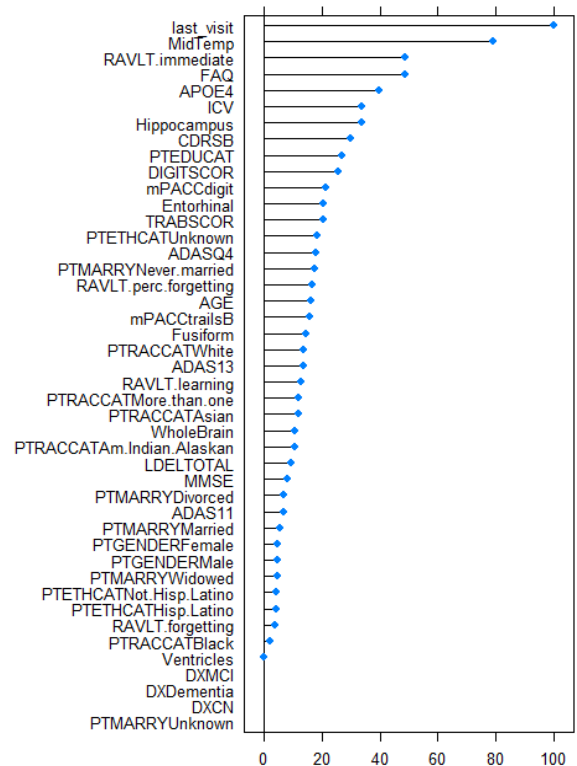

Fig. 2. Variable Importance for the MCI group from an Elastic Net model.

Of further interest is the differing variable importance measures for models in the two datasets. The best model in the CN_b group was CART, but this technique does not provide a natural way to calculate variable importance. However, we can look at the variable importance chart for a related model (Random Forest) and observe that the most important variables were neuropsychological tests such as the LDELTotal and the CDRSB. In comparison, the variable importance plot for the MCI_b group, derived from the best model (Elastic Net) demonstrates the most important variables to be the time to the last visit (measured in months), and a raft of neuroimaging metrics. This would seem to indicate that early deterioration (from CN to MCI/AD) indicators can be picked up by neuropsychological testing at baseline, but later deterioration would seem to rely more on changes detectable by neuroimaging. The former would support the work of Wang et al who used the ADNI repository to demonstrate good predictive power (AUC 0.83) in a model predicting, using baseline neuropsychological test results, those cognitively normal who would go on to develop MCI [12]. For the latter, there have been several studies indicating the effectiveness of using neuroimaging data to predict MCI deterioration e.g. [13]. However, our results would suggest that this data demonstrates a particular sensitivity for prediction in subjects with Mild Cognitive Impairment.

As mentioned previously, this work aims to contribute to an existing body of literature which addresses deterioration in dementia. This aspect of Alzheimer's research is challenging and multi-faceted. Indeed, a recent review of current literature concluded that the progression within dementia is heterogenous, both within and between persons, and originates from disease characteristics. The same paper recommends that possible risk factors should be evaluated at baseline, but also at intervals during disease progression [13]. To the first point, our current work suggests that heterogeneity exists, at the very least between the two groups described here. Such heterogeneity would indicate differing approaches may be appropriate with these groups, when selecting treatments or interventions. To the second point, our paper suggests a difference in the expression of risk factors at baseline, as compared to the subject's final visit. This does indeed support the conclusions of the review.

However, a limitation to this study that should be kept in mind is the relatively small sample size, particularly within the CN_b group. In this dataset the class imbalance was 90% CN to 10% MCI/AD. This impacted the overall stability of the models generated and, although we took steps to account for this, using the nested cross-validation and repeated nested cross-validation approaches, we cannot predict how the model may hold up when presented with a larger test set. The decision was made to only use data gathered through the ADNI1 study protocol and this significantly limited the sample size. This decision was reached because the diagnostic criteria differed markedly across the different ADNI studies, and therefore direct comparisons between the classifications would have been difficult. Thus, an obvious avenue to explore is to compare the results from this work with future models, built on the back of other ADNI studies, such as ADNI2 ADNIGO, or ADNI3.